\documentclass[12pt]{article}
\usepackage[utf8]{inputenc}
\usepackage{fullpage}

\usepackage{graphicx}
\usepackage{caption}
\usepackage{subcaption}
\usepackage{amsmath}
\usepackage[export]{adjustbox}
\usepackage{booktabs}
\usepackage{url}

\begin{document}
\title{A Brief Introduction to Automatic Differentiation for Machine Learning  \\
  }
    
\author{Davan Harrison\\University of California, Santa Cruz}
\date{June 9, 2019}
\maketitle

\begin{abstract}
Machine learning, and neural network models in particular, have been
improving the state of the art performance on many artificial intelligence
related tasks. Neural network models are typically implemented using 
frameworks that perform gradient based 
optimization methods to fit a model to a dataset.
These frameworks use a technique of calculating derivatives called automatic
differentiation (AD) which removes the burden of performing derivative 
calculations from the model designer. In this report we describe AD, its motivations, and different implementation approaches. We briefly describe 
dataflow programming as it relates to AD. Lastly, we present example 
programs that are implemented with Tensorflow and PyTorch, which 
are two commonly used AD frameworks.
\end{abstract}

\section{Introduction}
Deep learning is modeling technique which involves constructing large and complicated neural network models to perform tasks in a variety of artificial intelligence related fields, such as Computer Vision and Natural Language Processing. A neural network model uses complicated mathematical expressions that operate on matrix or tensor variables. Typically, neural network models are  fit to a dataset using gradient based optimization methods that involve repeated computation of derivatives.
In the recent years a number of neural network modeling frameworks have been developed, such as Theano 
\cite{bergstra2011theano}, Tensorflow \cite{abadi2016tensorflow}, and  PyTorch \cite{paszke2017pytorch}. These frameworks employ automatic differentiation (AD), which is a method of computing derivatives of numeric functions that are defined programmatically. AD allows for efficient and accurate evaluation of derivatives while at the same time removing the burden of performing derivative calculations from the model designer. 

In this report we discuss AD and its different forms. Also, we will discuss dataflow programming which is a programming paradigm relevant to AD.  Lastly, we will discuss the Tensorflow \cite{abadi2016tensorflow} and  PyTorch \cite{paszke2017pytorch} machine learning frameworks, and present example programs that are implemented using these frameworks. 

\section{Dataflow Programming}
\begin{figure}[]
    \begin{subfigure}[]{0.65\textwidth}
    	\centering
        \includegraphics[width=\linewidth]{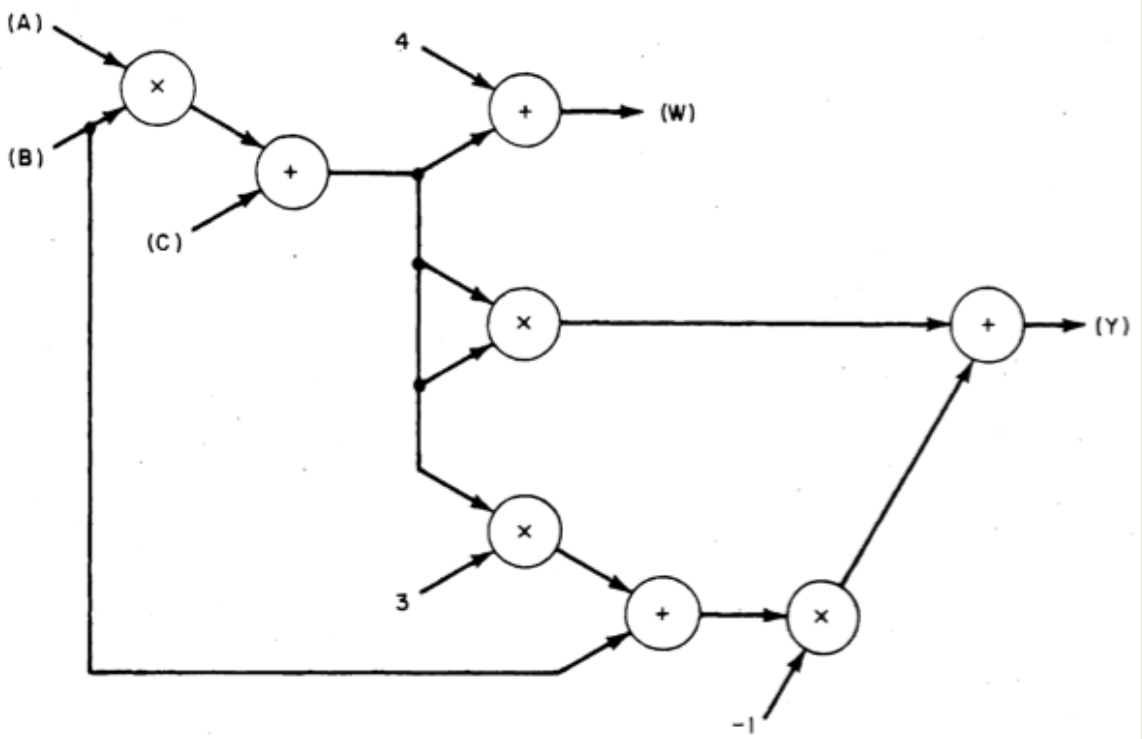}
        \caption{dataflow graph.}
    \end{subfigure}
    \begin{subfigure}[]{0.25\textwidth}
        \begin{small}
        \begin{align*}
            Z &:= A \times B + C \\
            W &:= Z + 4 \\
            Y &:= Z^2 - (3Z + B) 
        \end{align*}
        \end{small}
        \caption{written statement.}
    \end{subfigure}
    \caption{An example graphical representation of a written arithmetic calculation.}
    \label{fig:dataflow-graph}
\end{figure}

Dataflow programming is a programming paradigm where programs are represented as directed graphs \cite{sutherland1966line, sousa2012dataflow}. The dataflow programming paradigm has a concept called an 
executable block \cite{sousa2012dataflow}.
Thus, under this paradigm, the execution model gets represented as a graph where executable blocks are the nodes, and the edges between nodes represent data dependencies. The dataflow graph of an entire program will consist of at least one source node, at least one end node, and a set of data processing nodes. All the nodes are connected by directed data dependency edges. The nodes of the graph are primitive instructions such as arithmetic or comparison operations, and can have multiple data inputs (directed edges flowing toward a node) and outputs (directed edges flowing out of a node)\cite{johnston2004advances}. Once all of a node's input edges contain data, the node becomes \textit{fireable}, and will be executed at some future time \cite{johnston2004advances}. The node becomes dormant (waiting to become fireable) once the computation is finished, and the node's result gets passed on by its outward edges to other nodes that depend on the result. Figure~\ref{fig:dataflow-graph} shows an example sequential program given in \cite{sutherland1966line, sousa2012dataflow} and its dataflow representation.  

The benefit of the dataflow paradigm is that it easily supports concurrent execution of instructions because in the cases that multiple nodes become fireable at the same time, then they can all be executed in parallel. 

Many AD frameworks utilize the dataflow programming paradigm to construct programs in the form of  \textit{computation graphs} that can be operated on using the backpropogation algorithm.

\section{Automatic Differentiation}
In this section we will discuss AD and its motivation, and then describe the forward and backward modes of AD.  
The section will close with a brief discussion of two commonly used implementations of AD: operator overloading (OO), and source transformation (ST) \cite{van2018automatic}.

\subsection{Why AD?} Machine learning frequently involves computing and evaluating derivatives in order to optimize an objective function. There are four methods of programmatically computing derivatives: computing them by hand and coding them up, approximating derivatives with numerical differentiation, symbolic differentiation as in Maple or Mathematica, and lastly \textit{automatic differentiation}. In a machine learning setting, AD is preferred over the other three methods due to a number of limiting factors. Manual differentiation is time consuming and prone to error, while numerical differentiation is inaccurate due to truncation and rounding errors \cite{baydin2018automatic}. Symbolic differentiation suffers from "expression swell" and imposes restrictions on the programmer, such as limiting use of branching, loops, and recursion \cite{baydin2018automatic}. 

AD is different from numerical and symbolic differentiation but it has some overlap. AD is partly symbolic and partly numerical because it uses symbolic differentiation rules to produce numerical values of derivatives\cite{baydin2018automatic}. An AD system computes derivatives by accumulating values produced in the process of executing program instructions. Then numerical derivative evaluations are produced, as opposed to derivative expressions. The strengths of AD over the other methods is that it allows for accurate derivative computations, is computationally efficient, and requires minimal code changes in order to be used \cite{baydin2018automatic}.  

\subsection{Modes of AD}
When AD is applied to a program, the program is augmented so that it also computes derivatives alongside its standard computations. This is achieved by decomposing a program into its primitive operations, (e.g., binary arithmetic operations, transcendental functions, etc.) that have known derivatives \cite{baydin2018automatic, van2018automatic}. Then the derivative of the original program structure is calculated by using the chain rule from calculus to combine the derivatives of the program's primitive constituents. This process can be represented as an \textit{evaluation trace}\footnote{also known as a Wangert list} of a program \cite{baydin2018automatic}.

AD has two modes, forward mode (left to right) and reverse mode (right to left), that each correspond to the direction in which the chain rule is evaluated \cite{baydin2018automatic, van2018automatic}. In forward mode, intermediate computations of the original program and computations of derivatives are calculated alongside each other using a single monolithic pass over the program, or by making multiple passes (one pass for each input variable). Forward mode has constant memory requirements and computation complexity depends on the number of inputs \cite{van2018automatic}. 

In reverse mode, the chain rule is evaluated in the reverse order of the original program, sometimes called the \textit{primal} program. This is achieved by constructing a second program called an \textit{adjoint} program which has a control flow in reverse order to that of the original program \cite{van2018automatic}. Program execution is broken into two separate phases. In the first phase, the primal program is executed. Then, in the second phase, the adjoint program is executed to calculate the gradients by starting with the output of the primal program, continuing through the intermediate variables, and ending with the primal program's inputs\cite{van2018automatic}.
All the primal program's intermediate variables need to be saved until the adjoint program completes it execution because they are used for computing the gradients. Therefore, reverse mode has memory requirements that grow with the number of intermediate variables, and has computation complexity that grows with the number of outputs \cite{van2018automatic}. In machine learning, reverse mode AD is more frequently used due to the fact that many tasks in machine learning have few inputs and involve gradient based optimization of a scalar error value, which cause it to be more efficient than forward mode AD. Reverse mode AD corresponds to the backpropagation algorithm \cite{baydin2018automatic, van2018automatic}.

\begin{figure}[]
    	\centering
        \includegraphics[width=0.75\linewidth]{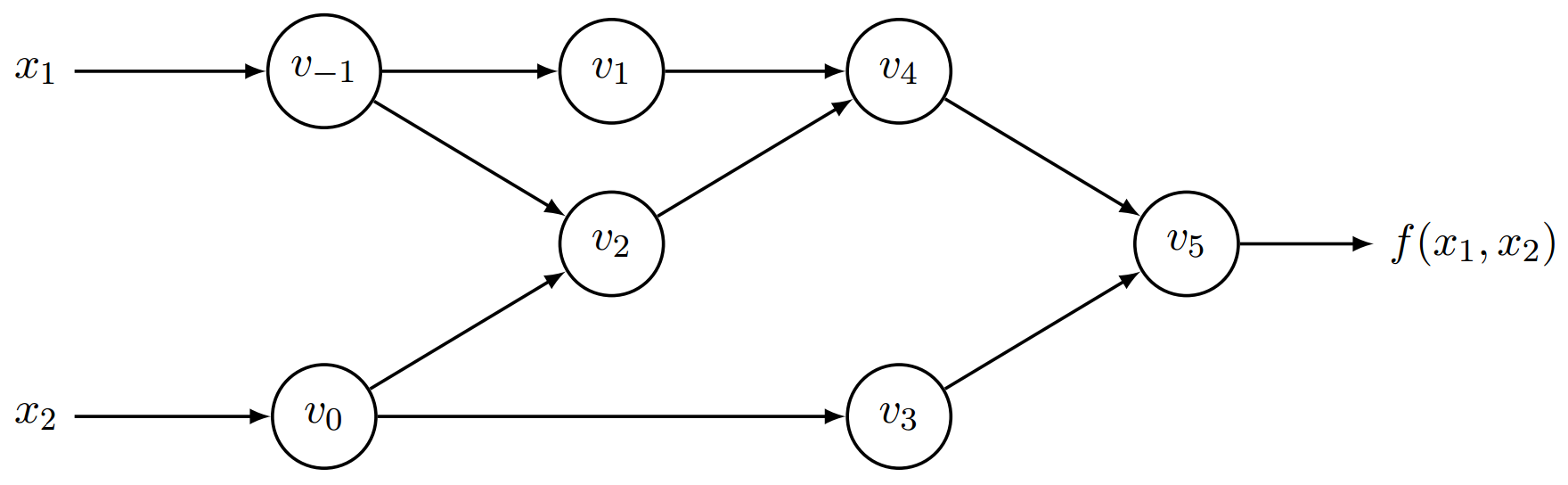}
        \caption{Computation graph for $y = f(x_1, x_2) = \text{ln}(x_1) + x_1x_2 - \text{sin}(x_2)$.}
        \label{fig:computation-graph}
\end{figure}

Consider this example computation whose computation graph is shown in Figure~\ref{fig:computation-graph}:
\begin{equation}
\label{func:f}
    y = f(x_1, x_2) = \text{ln}(x_1) + x_1x_2 - \text{sin}(x_2) \: ,
\end{equation}
and is an example taken from \cite{van2018automatic}. Table~\ref{table:reverse-mode-evaluation} shows an evaluation trace for Equation \ref{func:f} when using reverse mode AD. In Figure~\ref{fig:computation-graph} and Table~\ref{table:reverse-mode-evaluation}, $v_i$ for $1 \leq i$ represent intermediate variables, and $v_i$ for $i \leq 0$ represent input variables. The output variable is represented by $y$.

\begin{table}[]
\centering
\begin{tabular}{ll|lll} \toprule
\multicolumn{2}{c}{Forward Primal Trace} & \multicolumn{3}{c}{Reverse Adjoint Trace} 
\\ \midrule
$v_{-1}=x_1$ & $=2$                         &$x'_1=v'_{-1}$  & & $=5.5$  \\
$v_0=x_2$ & $=5$                            &$x'_2=v'_{0}$  & & $=1.716$  \\ \midrule

$ v_1=\text{ln}(v_{-1})$ &  $=\text{ln}(2)$ 
&$v'_{-1}=v'_{-1} + v'_1\frac{\partial v_1}{\partial v_{-1}}$  &$=v'_{-1} + v'_1/v_{-1}$ & $=5.5$  \\

$ v_2=v_{-1}\times v_0$ &  $=2\times5$      
&$v'_0=v'_0 + v'_2\frac{\partial v_2}{\partial v_{0}}$  &$=v'_{0} + v'_2 \times v_{-1}$ & $=1.716$  \\

  &                                         
  &$v'_{-1}=v'_2 \frac{\partial v_2}{\partial v_{-1}}$  &$=v'_2 \times v_0$ & $=5$  \\

$ v_3=\text{sin}(v_0)$ &  $=\text{sin}(5)$  
&$v'_0=v'_3\frac{\partial v_3}{\partial v_{0}}$  &$=v'3 \times \text{cos}(v_0)$ & $=-0.284$  \\

$ v_4=v_1+v_2$ &  $=0.693 + 10$             
&$v'_2=v'_4\frac{\partial v_4}{\partial v_{2}}$  &$=v'_4 \times 1$ & $=1$  \\

  &                                         
  &$v'_{1}=v'_4\frac{\partial v_4}{\partial v_{1}}$  &$=v'_4 \times 1$ & $=1$  \\

$ v_5=v_4-v_3$ &  $=10.693+0.959$           
&$v'_3=v'_5\frac{\partial v_5}{\partial v_{3}}$  &$=v'_5 \times (-1)$ & $=-1$  \\ 

  &                                         
  &$v'_{4}=v'_5\frac{\partial v_5}{\partial v_{4}}$  &$=v'_5 \times 1$ & $=1$  \\ \hline

$ y=v_5$ &  $=11.652$                       &$v'_5=y'$  &$=1$ &      \\ \bottomrule
\end{tabular}
    \caption{Evaluation trace of $y = f(x_1, x_2) = \text{ln}(x_1) + x_1x_2 - \text{sin}(x_2)$ with $x_1 = 2$ and $x_2 = 5$ using revers mode AD.}
    \label{table:reverse-mode-evaluation}
\end{table}

\subsection{Operator overloading}
In the operator overloading (OO) approach to AD, the adjoint program is dynamically constructed upon execution of the primal program.
The definitions of primitive operations of the host programming language are overloaded so that they perform additional tasks that facilitate the evaluation trace \cite{van2018automatic}. As each operation of the primal program is executed, the inputs and results are saved on a "tape" in order to preserve intermediate variables for later use by the adjoint program.  Often, the tape takes the form of a graph that resembles the computation graphs used in dataflow programming. After the forward execution of the program, calculation of derivatives is achieved by processing the tape in the reverse direction \cite{van2018automatic}.

OO has several strengths and weaknesses. The backpropogation algorithm is simple to implement using the OO method  \cite{van2018automatic}. Also, it supports ease of use by allowing users to implement models as a regular program in the host language, thereby facilitating arbitrary control flow statements and promoting intuitive debugging \cite{baydin2018automatic}. On the other hand, the OO method incurs a runtime cost because the adjoint program is constructed at each execution of the primal program. Furthermore, it does not allow for static optimization of the adjoint program.

\subsection{Source Transformation}

In OO, the adjoint program is dynamically constructed at each execution and is defined as an implicit property of the primal program, but this is not case in the source transformation approach (ST). In ST, the user uses a domain specific mini language to define a computation graph that explicitly specifies both the primal and adjoint programs \cite{baydin2018automatic}. Then, during execution, the framework interprets the program with different inputs while keeping the computation graph fixed. Similar to OO, a tape can be used in the ST approach to save the intermediate variables during execution, or the intermediate variables can be saved directly on the computation graph.

A benefit of the ST approach is that it only builds the adjoint program once, and it allows for ahead of time optimization of the computation graph structure since the adjoint program is known during compilation \cite{van2018automatic, baydin2018automatic}. Furthermore, with ST, the framework is aware of the entire computation graph structure ahead of time, which makes it easier to take advantage of the parallelization opportunities provided by computation graphs and the dataflow programming paradigm. On the down side, this method can impose limitations on the user, such as limiting use of loops, recursion, and higher order functions \cite{van2018automatic}. In addition, programs using ST can be difficult to debug and have an un-intuitive control flow \cite{baydin2018automatic}.

\section{AD in PyTorch}
\begin{figure}[]
    \begin{subfigure}[]{0.6\textwidth}
    	\centering
        \includegraphics[width=\linewidth]{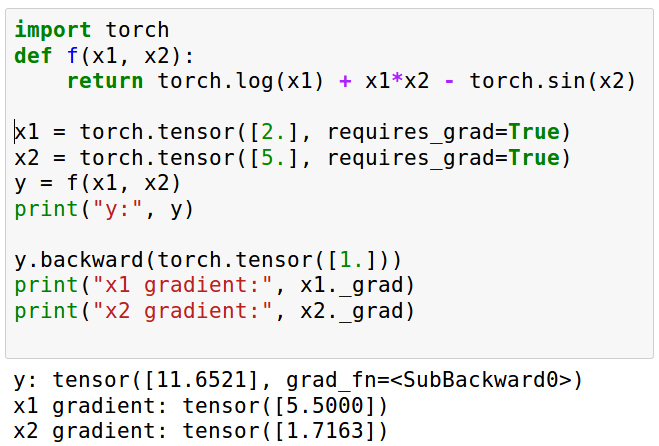}
        \caption{Python code.}
    \end{subfigure}
    \begin{subfigure}[]{0.35\textwidth}
    	\centering
        \includegraphics[width=\linewidth]{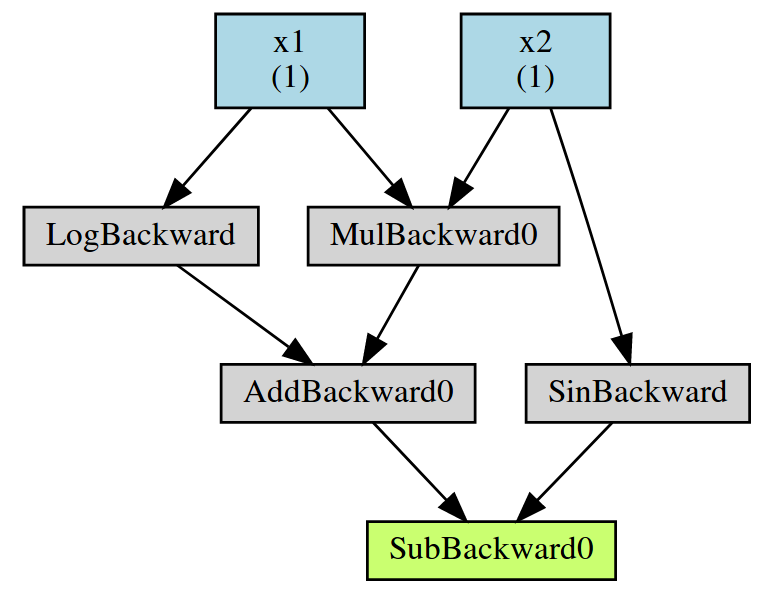}
        \caption{computation graph.}
    \end{subfigure}
    \caption{A PyTorch program that implements Equation~\ref{func:f} and the computation graph produced by PyTorch while executing the program.}
    \label{fig:pytorch-example}
\end{figure}

The PyTorch framework uses the OO approach to implement AD \cite{paszke2017pytorch}. PyTorch is a library for the Python programming language and allows users to use any Python features they want \cite{paszke2017pytorch}. Primitive instructions are overloaded so that they perform an evaluation trace and save intermediate variables, but the framework limits its use of tapes. Instead, the intermediate results record only the subset of the computation graph that relates to their computation which supports parallel execution and allows sections of the computation graph to be quickly freed from memory once they are no longer needed \cite{paszke2017pytorch}. Even though PyTorch is a Python library, most of it is implemented in C++ to improve run-time efficiency. Figure~\ref{fig:pytorch-example} shows a program that uses the PyTorch framework to implement the computation of Equation~\ref{func:f}, as well the computation graph that is created during the execution of the program.

\section{AD in Tensorflow}
\begin{figure}[]
    \begin{subfigure}[]{0.6\textwidth}
    	\centering
        \includegraphics[width=\linewidth]{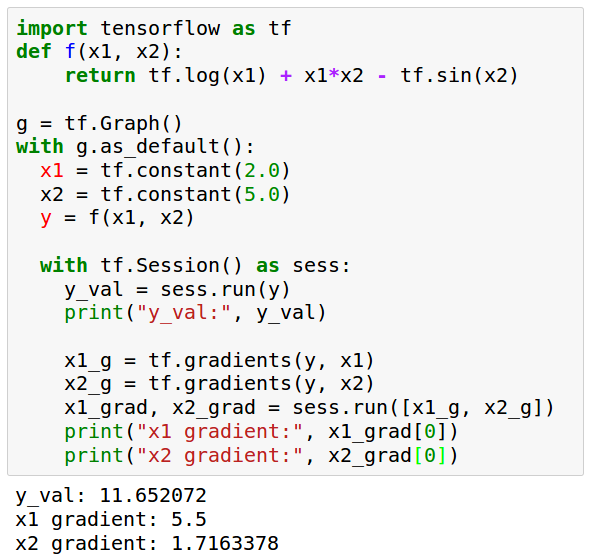}
        \caption{Python code.}
    \end{subfigure}
    \begin{subfigure}[]{0.35\textwidth}
    	\centering
        \includegraphics[width=\linewidth]{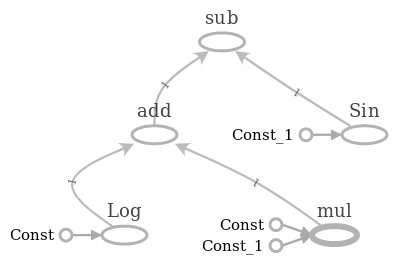}
        \caption{computation graph. Input variables are named with \textit{Const}.}
    \end{subfigure}
    \caption{A Tensorflow program that implements Equation~\ref{func:f} and a graphical representation of the computation graph constructed by Tensorflow.}
    \label{fig:tensorflow-example}
\end{figure}

The Tensorflow framework\cite{abadi2016tensorflow} follows the source transformation approach where the user explicitly constructs a computation graph.
A Tensorflow program that implements the computations of Equation~\ref{func:f} is shown in Figure~\ref{fig:tensorflow-example}. Note that the computation graph is explicitly referenced. Nodes are added to the graph by accessing it as a global variable as part of the Tensorflow instructions.   

Tensorflow programs get compiled using its XLA\footnote{\url{www.tensorflow.org/xla}} linear algebra compiler which optimizes computations. Unlike PyTorch, Tensorflow is language-independent. So computation graphs can be implemented in Python and compiled and saved using an intermediate representation. Then they can be loaded and run by a Tensorflow framework of a different language, such as C++, or Javascript.

\section{Conclusion}
In this report we describe automatic differentiation and its motivations. We discuss the forward and reverse modes of AD, and focus on the reverse mode because it is typically used in machine learning. Two approaches to reverse mode AD are described. Then, we present examples of similar programs implemented using two AD frameworks, namely, Tensorflow and PyTorch. Along the way we briefly discuss the dataflow programming paradigm which is heavily used by many AD frameworks, including Tensorflow and PyTorch.

\bibliographystyle{acm}
\bibliography{mybib}
\end{document}